\documentclass[11pt,a4paper]{article}
\usepackage[hyperref]{arxiv}
\usepackage{times}
\usepackage{latexsym}

\usepackage{microtype}

\aclfinalcopy 
\def\aclpaperid{3196} 

\usepackage[pdftex]{graphicx}
\usepackage{CJKutf8}
\usepackage{amsmath} 
\usepackage{amssymb} 
\usepackage{bm}
\usepackage{booktabs} 
\usepackage{multirow} 
\usepackage{color} 
\usepackage{subcaption} 
\usepackage{comment}

\newcommand{\figref}[1]{Figure~\ref{fig:#1}}

\newcommand{\tabref}[1]{Table~\ref{tab:#1}}

\newcommand{\ssecref}[1]{\S~\ref{subsec:#1}}

\newcommand{\argmin}{\mathop{\mathrm{argmin}}\limits}
\newcommand{\shoetsu}[1]{{#1}}
\newcommand{\yn}[1]{{#1}}
\newcommand{\jsaku}[1]{{#1}}

\title{Vocabulary Adaptation 
for 
Domain Adaptation\\
in Neural Machine Translation}

\author{
  Shoetsu Sato\quad 
  Jin Sakuma \\
  The University of Tokyo\\
  \And
  Naoki Yoshinaga\quad
  Masashi Toyoda\\
  Institute of Industrial Science,\\ 
  the University of Tokyo\\
  \AND
  Masaru Kitsuregawa \\
  Institute of Industrial Science,
  the University of Tokyo\\
  National Institute of Informatics \\ 
  \AND
  \\*[-19pt]
  \texttt{\{shoetsu,jsakuma,ynaga,toyoda,kitsure\}@tkl.iis.u-tokyo.ac.jp}
  }

\begin{document}
\maketitle
\begin{abstract}
\yn{Neural network methods 
exhibit strong performance only in a few 
resource-rich domains. 
Practitioners therefore employ domain adaptation from resource-rich domains that are, in most cases, distant from the target domain.}
Domain adaptation between distant domains (\textit{e.g.}, movie subtitles and research papers), however, cannot be performed effectively due to 
mismatches in vocabulary; 
it will encounter many 
domain-specific words (\textit{e.g.}, ``\textit{angstrom}'') and words whose meanings shift across domains (\textit{e.g.}, ``\textit{conductor}'').
In this study, aiming to solve these vocabulary mismatches in 
domain adaptation for neural machine translation (\textsc{nmt}), we propose \textit{vocabulary adaptation}, 
a simple method for effective fine-tuning that adapts embedding layers in a given pre-trained \textsc{nmt} model to the target domain.
Prior to fine-tuning, our method replaces the embedding layers of the \textsc{nmt} model by projecting general word embeddings induced from monolingual data in a target domain onto a source-domain embedding space.
Experimental results 
indicate that our method improves the performance of conventional fine-tuning by 3.86 and 3.28 \textsc{bleu} points in En$\rightarrow$Ja  and De$\rightarrow$En translation, respectively.
\end{abstract}


\section{Introduction}\label{sec:introduction}

\begin{figure}[t]
\centering
\includegraphics[width=0.99\linewidth,clip]{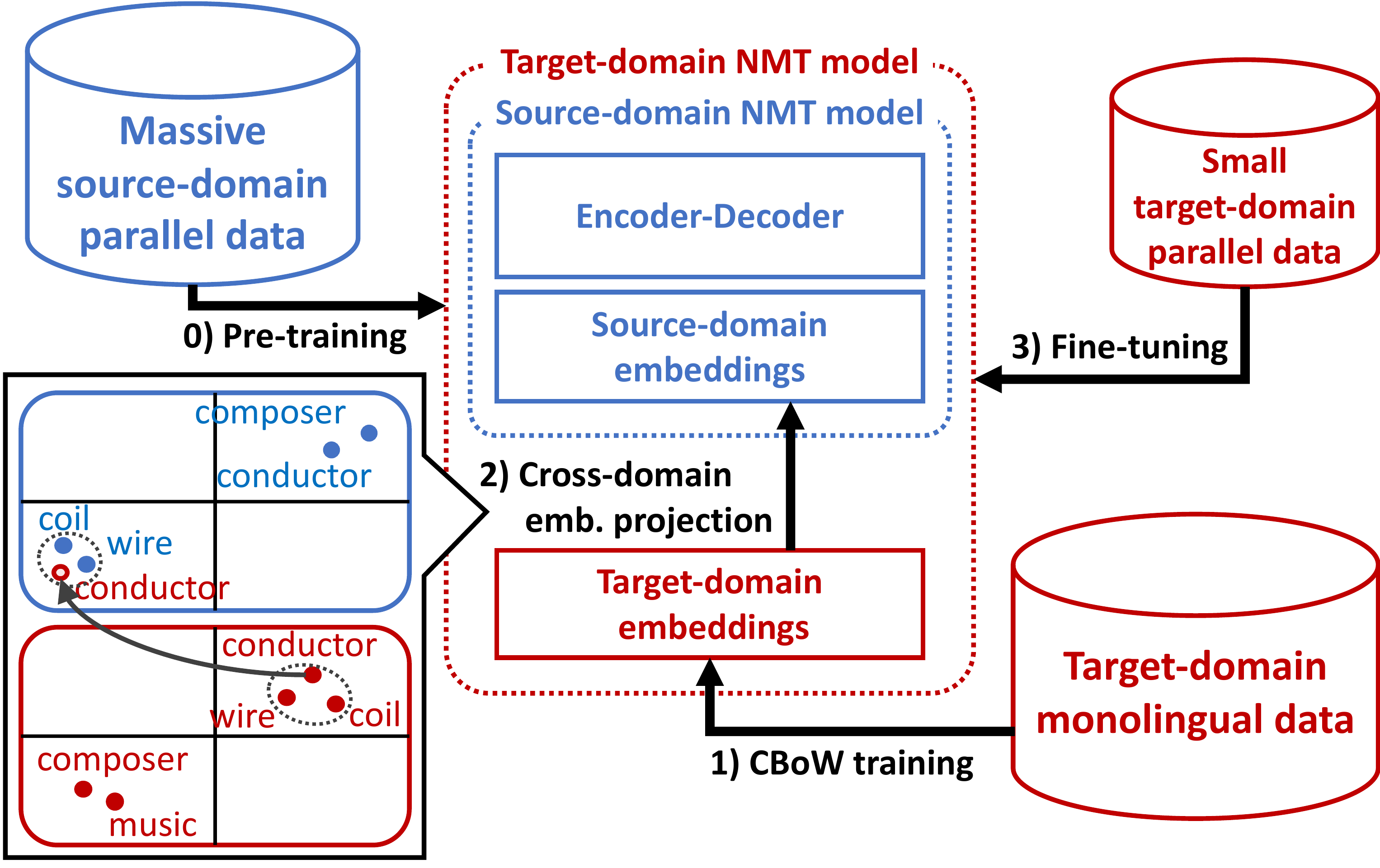}
\caption{Vocabulary adaptation for 
domain adaptation in NMT
using cross-domain embedding projection.}
\label{fig:overview}
\end{figure}

The performance of 
neural machine translation (\textsc{nmt}) models remarkably drops in domains different from the training data~\cite{koehn-2017-six}. 
Since a massive amount of parallel data is available only in a limited number of domains, 
domain adaptation 
is often required to employ \textsc{nmt} in practical applications.
Researchers have therefore developed fine-tuning, a dominant approach for this problem~\cite{luong2015stanford,DBLP:journals/corr/FreitagA16,chu-etal-2017-empirical,thompson-etal-2018-freezing,khayrallah-etal-2018-regularized,bapna-firat-2019-simple} (\S~\ref{sec:related-work}).
Assuming a massive amount of source-domain and small amount of target-domain parallel data, 
fine-tuning adjusts the parameters of a model pre-trained in the source-domain to the target domain. 

\yn{However, in fine-tuning, 
inheriting 
the embedding layers of the model pre-trained in the source domain causes vocabulary mismatches;}
namely, a model can handle neither domain-specific words that are \jsaku{not covered}
by a small amount of target-domain parallel data (\textit{unknown words})
nor words that have different meanings across domains (\textit{semantic shift}).
Moreover, 
adopting the standard subword tokenization~\cite{sennrich-etal-2016-neural, kudo-2018-subword}  
accelerates the semantic shift.
Target-domain-specific words are often finely decomposed into source-domain subwords (\textit{e.g.}, ``alloy'' $\rightarrow$ ``\_all''
+ ``o'' + ``y''), which 
introduces improper subword meanings
and hinders adaptation (\tabref{output-examples-trans} in \S~\ref{sec:results}). 

To resolve these vocabulary-mismatch problems in 
domain adaptation, we propose \textit{vocabulary adaptation}  (Figure~\ref{fig:overview}), a method of directly adapting the vocabulary (and embedding layers) of a pre-trained \textsc{nmt} model to a target domain, to perform effective fine-tuning (\S~\ref{sec:proposal}). 
Given an \textsc{nmt} model pre-trained in a source domain, we first induce 
a wide coverage of target-domain word embeddings from target-domain  monolingual data. 
We then fit the obtained target-domain word embeddings to the embedding space of the pre-trained \textsc{nmt} model by inducing a cross-domain
projection from the target-domain embedding space to the source-domain embedding space. To perform this cross-domain embedding projection, we explore two methods: 
cross-lingual~\cite{xing2015} and cross-task embedding projection~\cite{sakuma2019}.

We evaluate fine-tuning with the proposed vocabulary adaptation for two domain pairs:  \jsaku{1)} from \textbf{JESC}~\cite{pryzant_jesc_2018} to \textbf{ASPEC}~\cite{nakazawa2016aspec} for English to Japanese translation (En$\rightarrow$Ja) and 
 \jsaku{2)} from the \textbf{IT} domain to \textbf{Law} domain~\yn{\cite{koehn-2017-six}}
 for German to English translation (De$\rightarrow$En).
Experimental results demonstrate that our 
vocabulary adaptation improves the \textsc{bleu} scores~\cite{papineni-etal-2002-bleu} of fine-tuning~\cite{luong2015stanford}
by 3.86 points (21.45 to 25.31) for En$\rightarrow$Ja and 3.28 points (24.59 to 27.87) for De$\rightarrow$En\jsaku{~(\S~\ref{sec:results})}.
Moreover, it shows further improvements 
when combined with back-translation~\cite{sennrich-etal-2016-improving}.

The contributions of this paper are as follows.
\begin{itemize}
\item We empirically confirmed that \textbf{vocabulary mismatches 
hindered domain adaptation}. 
\item We established \textbf{an effective, model-free 
fine-tuning
for \textsc{nmt}} that adapts the vocabulary
of a pre-trained 
model to a target domain.
\item We showed that \textbf{vocabulary adaptation exhibited additive improvements over back-translation} that uses monolingual corpora.
\end{itemize}

\section{Related Work}\label{sec:related-work}
In this section, we first review
two approaches to \yn{supervised} domain adaptation in \textsc{nmt}: multi-domain learning and fine-tuning. 
We then introduce unsupervised domain adaptation using target-domain monolingual data and 
approaches to
unknown word problems in \textsc{nmt}.

\smallskip\noindent\textbf{Multi-domain learning}
induces an \textsc{nmt} model from parallel data in both source and target domains~\cite{kobus-etal-2017-domain,wang-etal-2017-instance,britz2017effective}. 
Since this approach requires training with a massive amount of source-domain parallel data,
the training cost becomes problematic when we perform adaptation to many target domains.

\smallskip\noindent\textbf{Fine-tuning} 
(or continued learning)
is a standard domain adaptation method in \textsc{nmt}.
Given an \textsc{nmt} model pre-trained with a massive amount of source-domain parallel data,
it continues the training of this pre-trained model with a small amount of target-domain parallel data~\cite{luong2015stanford,chu-etal-2017-empirical,thompson-etal-2018-freezing,bapna-firat-2019-simple,gu-etal-2019-improving}. 
Due to the small cost of training, research trends have shifted to fine-tuning from multi-domain learning.
Recent studies focus on 
model architectures, training objectives, and strategies in training.
Meanwhile, no attempts have been made 
to resolve the vocabulary mismatch problem in domain adaptation.


\smallskip\noindent\textbf{Unsupervised domain adaptation} exploits target-domain monolingual data to train a language model 
to support 
the model's decoder in
generating natural sentences in a target domain~\cite{DBLP:journals/corr/GulcehreFXCBLBS15,domhan-hieber-2017-using}. 
Data augmentation using back-translation
~\cite{sennrich-etal-2016-improving, hu-etal-2019-domain-adaptation} 
is another approach to using target-domain monolingual data. 

These approaches can partly address the problem of semantic shift.
However, it is possible that the source-domain encoder will fail to handle target-domain-specific words. In such cases, a decoder with the target-domain language model becomes less helpful in the former approach, and the generated pseudo-parallel corpus has low-quality sentences on the encoder side in the latter approach.

\smallskip\noindent\textbf{Handling unknown words} has been extensively studied for \textsc{nmt} since the vocabulary size of an \textsc{nmt} model is limited due to practical requirements (\textit{e.g.}, GPU memory) ~\cite{jean-etal-2015-using,luong-etal-2015-addressing}. 
The current standard approach to the unknown word problem is to use token units shorter than words such as characters~\cite{ling2015character,luong-manning-2016-achieving} and subwords~\cite{sennrich-etal-2016-neural,kudo-2018-subword} to handle rare words as a sequence of known \yn{tokens}.
However, more drastic semantic shifts will occur for
characters or subwords
than for words because they are shorter than words and naturally ambiguous.




\shoetsu{Besides these studies mentioned above, \citet{aji-etal-2020-neural} reported that transferring embeddings and vocabulary mismatches between parent and child models significantly affected the performance of models also in cross-lingual transfer learning. 
} 

In this study, we aim to provide pre-trained \textsc{nmt} models with functionality that directly handles both target-domain-specific unknown words and semantic shifts 
by exploiting cross-domain embeddings learned from target-domain 
data. 

\section{Vocabulary Adaptation for Domain Adaptation in \textsc{nmt}}
\label{sec:proposal}
As we have discussed 
(\S~\ref{sec:introduction}), 
vocabulary mismatches between source and target domains are the important challenge in domain adaptation for \textsc{nmt}\@. 
This section proposes fine-tuning-based methods of directly resolving this problem. 
Although our methods are applicable to any \textsc{nmt} model with embedding layers, we assume here subword-based encoder-decoder models~\cite{Bahdanau2015a,vaswani2017attention} for clarity.




\subsection{Vocabulary Adaptation \yn{Prior to} Pre-training}\label{subsec:ft-tgtv}
One simple approach is to use target-domain vocabularies 
in pre-training. 
Specifically, we first construct vocabularies 
from \textit{target-domain} 
data for each language. 
We then pre-train an \textsc{nmt} model in a source domain with the target-domain vocabularies and embeddings. Finally, we fine-tune the pre-trained model with target-domain parallel data.

In this approach, however, 
employing the target-domain vocabularies 
\yn{will} hinder pre-training in the source domain. In addition, since the embeddings induced from the target-domain data are 
\yn{tuned}
to the source domain,
the problem of semantic shifts still remains  \yn{and will hinder fine-tuning}. 

\subsection{Vocabulary Adaptation \yn{Prior to} Fine-tuning}\label{subsec:vocabulary-adaptation}
Another approach is to replace 
the encoder's embeddings and the decoder's embeddings 
of the pre-trained \textsc{nmt} model with word embeddings induced from target-domain data before fine-tuning.
However, as in transplanting organs from a donor to a recipient, this causes rejection; the embedding space of a pre-trained model is irrelevant to the space of the target-domain word embeddings. 

We therefore project the target-domain word embeddings 
onto the embedding space of the pre-trained model in order to make the embeddings compatible with the pre-trained model (Figure~\ref{fig:overview} in \S~\ref{sec:introduction}). This approach is inspired by cross-lingual and cross-task word embeddings that bridge word embeddings across languages and tasks.

An overview of our proposed method is given as follows.

\smallskip\noindent\textbf{Step 1 (Inducing target-domain embeddings)} We 
    induce word embeddings from monolingual data in the target domain for each language. Although we can use any method for induction, we adopt Continuous Bag-of-Words (\textsc{cbow})~\cite{mikolov2013distributed} here since \textsc{cbow} is effective for initializing embeddings in \textsc{nmt}~\cite{neishi-etal-2017-bag}, \yn{which suggests embedding spaces of \textsc{cbow} and \textsc{nmt} are topologically similar.}
    
    \smallskip\noindent\textbf{Step 2 (Projecting embeddings across domains)} We project the target-domain embeddings of the source and target languages 
    into the embedding spaces of the pre-trained encoder and decoder, respectively, to obtain cross-domain embeddings~(\S~\ref{subsec:linearmap}, \S~\ref{subsec:LLM}).
    
    \smallskip\noindent\textbf{Step 3 (Fine-tuning)} We replace the vocabularies and 
    the embedding layers with the cross-domain embeddings and apply fine-tuning using the target-domain parallel data.

To induce cross-domain embedding projection, we regard the two domains as different languages/tasks and explore the use of methods for inducing cross-lingual~\cite{xing2015} and cross-task word embeddings~\cite{sakuma2019}.
In what follows, we explain each method.

\subsubsection{Vocabulary Adaptation by Linear Transformation}\label{subsec:linearmap}

The first method 
exploits an orthogonal linear transformation~\cite{xing2015} to obtain cross-lingual word embeddings.
We use 
subwords shared across two domains 
for inducing
an orthogonal linear transformation from the embeddings of the target domain to the embeddings of the source domain.
The obtained linear transformation is used to map all embeddings of the target domain to the embedding space of the source domain \yn{to address semantic shift across domains}. 

\subsubsection{Vocabulary Adaptation by Locally Linear Mapping}\label{subsec:LLM}
Due to the difference between the domains and
tasks (\textsc{cbow} and \textsc{nmt}) in inducing the embeddings, the linear transformation is likely to fail.
Thus, we employ a recent method for cross-task embedding projection called ``locally linear mapping'' (LLM)~\cite{sakuma2019}. 
\shoetsu{An overview is illustrated in \figref{overview} (lower left).}

\begin{figure}[t]
\centering
\includegraphics[width=\linewidth,clip]{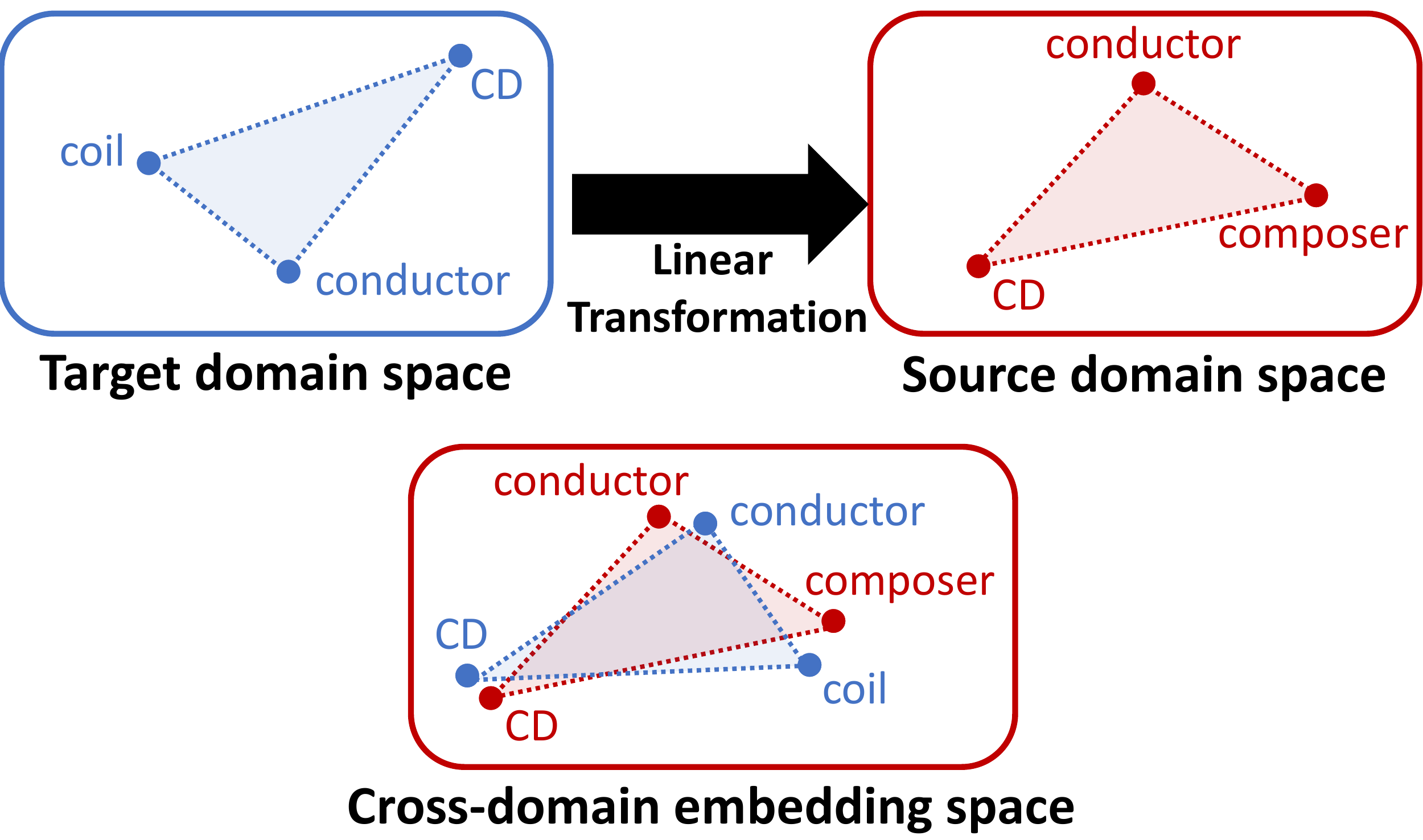}
\caption{\textbf{Unwanted} cross-domain projection by linear transformation due to difference of topology in vector-based embedding space: illustrative example.}
\label{fig:cross-domain-mapping}
\end{figure}

LLM 
learns a projection that preserves the local topology (positional relationships) of the original embeddings after mapping while disregarding the global topology.
This property of LLM is suited to our situation because the local topology is expected to be the same across the semantic spaces of two domains, while globally, they 
can be
significantly different 
due to semantic shift between domains
as illustrated in \figref{cross-domain-mapping}.

Here, we explain the essence of LLM\@.  Interested readers may consult
\citet{sakuma2019}
for details.
Suppose that $T^\textrm{LM}$ is the word embeddings of the target domain induced by a language model task, and $S^\textrm{NMT}$ is the word embeddings of the source domain induced by the translation task (the embedding layer of the pre-trained model).
We denote the vocabulary of $T^\textrm{LM}$ by $V_T$, the vocabulary of $S^\textrm{NMT}$ by $V_S$ and the vocabulary of words shared across both domains by $V_\textrm{shared}=V_T\cap V_S$\@.

Our goal is to produce embeddings $T^\textrm{NMT}$ with a vocabulary of $V_T$ in the embedding space of $S^\textrm{NMT}$.
We accomplish this by computing the $T^\textrm{NMT}$ that best preserves the local topology of $T^\textrm{LM}$ in the embedding space of $S^\textrm{NMT}$.
Concretely, for each word $w_i$ in $V_T$, we first take the $k$-nearest neighbors $\mathcal{N}(w_i)\subset V_\textrm{shared}$ in $T^\textrm{LM}$.
We use cosine similarity as the metric for the nearest neighbor search.

Second, we learn the local topology around $w_i$ by reconstructing $T^\textrm{LM}_{w_i}$ from the embeddings of its nearest neighbors as a weighted average.  For this purpose, we minimize the following objective:
\begin{align}
    \hat{\bm{\alpha}}_{i} = \argmin_{\bm{\alpha}_i} \left\|T^\textrm{LM}_{w_i}-\sum_{w_j\in\mathcal{N}(w_i)}\alpha_{ij}T^\textrm{LM}_{w_j}\right\|^2,
\end{align}
with the 
constraint of $\sum_{j}\alpha_{ij}=1$; 
the method of Lagrange multipliers gives the analytical solution.

We then compute the embedding $T^\textrm{NMT}_{w_i}$ that preserves the local topology by minimizing the following objective function:
\begin{align}
    T^\textrm{NMT} = \argmin_{T^\textrm{NMT}} \left\|T^\textrm{NMT}_{w_i}-\sum_{w_j\in\mathcal{N}(w_i)}\hat{\alpha}_{ij}S^\textrm{NMT}_{w_j}\right\|^2.
\end{align}
This optimization problem has the 
trivial solution:
\begin{align}
T^\textrm{NMT}_{w_i} = \sum_{w_j\in\mathcal{N}(w_i)}\hat{\alpha}_{ij}S^\textrm{NMT}_{w_j}.\label{eq:yspec}
\end{align}

Note that \yn{subwords} shared across domains will have different embeddings after projection ($T^\textrm{NMT}_{w}\neq S^\textrm{NMT}_w$ for $w\in V_\textrm{shared}$).  
This captures the semantic shift of \yn{subwords} across domains. We conduct a detailed analysis of this matter in \ssecref{analysis-emb}.

\section{Experimental Setup}\label{sec:experiments}
We conducted fine-tuning with our vocabulary adaptation for 
domain adaptation in En$\rightarrow$Ja and De$\rightarrow$En machine translation. 
In what follows, we describe the setup of our experiments.

\subsection{Datasets and Preprocessing}
We selected domain pairs to simulate a plausible situation where the target domain is specialized and similar source-domain parallel data is  not available.

For En$\rightarrow$Ja translation, we chose the Japanese-English Subtitle Corpus (JESC)~\cite{pryzant_jesc_2018}
 as the source domain and Asian Scientific Paper Excerpt Corpus (ASPEC)~\cite{nakazawa2016aspec}
  as the target domain.
JESC \yn{was} 
constructed from subtitles of movies and TV shows, while ASPEC was constructed from \yn{abstracts of} scientific papers.  
These domains are substantially distant, and 
ASPEC 
contains many 
technical terms that are unknown in the JESC domain.
We followed the official splitting of training, development, and test sets, except that the last 1,000,000 sentence pairs were omitted in the training set of the ASPEC corpus as they contain low-quality translations.

For De$\rightarrow$En translation, we adopted the dataset constructed by \citet{koehn-2017-six} from the OPUS corpus~\cite{tiedemann-2012-parallel}. This dataset includes multiple domains that are distant from each other and is suitable for experiments on 
realistic domain adaptation.
We chose the \textbf{IT} domain and the \textbf{Law} domain from the dataset as the source and target domain, respectively. 
We followed the same splitting 
of training, development, and test sets
as \citet{koehn-2017-six}.

\begin{table}[!t]
\footnotesize
\centering
\begin{tabular}{@{\,}l@{\quad}l@{}r@{\,}c@{\,}r@{\,}}
\toprule
 \textbf{En$\rightarrow$Ja}  & & \multicolumn{1}{c@{\,}}{\textbf{JESC}} & $\rightarrow$  & \multicolumn{1}{@{\,}c@{\,}}{\textbf{ASPEC}}   \\ 
\midrule 
 \multirow{3}{*}{\# examples}    
          & training (all)    & 2,797,388 & & 2,000,000     \\
          & development \,\, & 2,000     & & 1,790  \\
          & testing     & -         & & 1,812   \\ 
\midrule
\multicolumn{2}{l}{\# distinct words (En)} & 161,695 & & 637,377  \\
\multicolumn{2}{l}{\# distinct words (Ja)} & 169,649 & & 384,077 \\
\multicolumn{2}{l}{\# shared words (En)} & \multicolumn{3}{l@{\,}}{\hfill \mbox{46,950\ \ \ (7.4\% in ASPEC)}} \\
\multicolumn{2}{l}{\# shared words (Ja)} & \multicolumn{3}{l@{\,}}{\hfill \mbox{50,003 (13.0\% in ASPEC)}} \\

\toprule
\textbf{De$\rightarrow$En}  & & \multicolumn{1}{c@{\,}}{\textbf{IT}} & $\rightarrow$  & \multicolumn{1}{@{\,}c@{\,}}{\textbf{Law (Acquis)}}   \\ 
\midrule 
 \multirow{3}{*}{\# examples}    
          & training (all)    & 337,817 & & 715,372    \\
          & development       & 2,526   & & 2,000 \\
          & testing           & -       & & 2,000  \\ 
\midrule
\multicolumn{2}{l}{\# distinct words (De)} & 140,508 & &189,084   \\
\multicolumn{2}{l}{\# distinct words (En)} & 70,650 & & 92,316   \\
\multicolumn{2}{l}{\# shared words (De)} & \multicolumn{3}{l@{\,}}{\hfill \mbox{21,912} (11.6 \% in Law)} \\
\multicolumn{2}{l}{\# shared words (En)} & \multicolumn{3}{l@{\,}}{\hfill \mbox{17,165 (18.6 \% in Law)}} \\

\bottomrule
\end{tabular}
\caption{Statistics of source and target parallel corpus. \#distinct/shared words are counted in training sets.}
\label{tab:data-stat}

\end{table}


\paragraph{Preprocessing}
As preprocessing 
for the En$\rightarrow$Ja datasets, we
first tokenized the parallel data using the Moses toolkit (v4.0)\footnote{\url{https://github.com/moses-smt/mosesdecoder}} for English sentences and KyTea (v0.4.2)\footnote{\url{http://www.phontron.com/kytea}} for Japanese sentences. We then truecased the English sentences by using the script in the Moses toolkit.
As for the De$\rightarrow$En datasets, we used the same tokenization and truecasing as~\citet{koehn-2017-six}. 
The statistics of the datasets are listed in \tabref{data-stat}.

We applied SentencePiece (v0.1.83)\footnote{\url{https://github.com/google/sentencepiece}}~\cite{kudo-richardson-2018-sentencepiece} trained from the monolingual data in each domain to the tokenized datasets. The number of subwords was 16,000 for all languages. 
In the training of SentencePiece, we did not concatenate the input language and output language to maximize the portability of the pre-trained model.

From each of the preprocessed datasets, 
we used 1) 100,000 randomly sampled sentence pairs or 2) all sentence pairs in the training set for training in the target domain. This was for evaluating models in both cases where we have a small/large target-domain dataset.

\shoetsu{
To prepare reproducible target-domain monolingual data,
we shuffled and divided all sentence pairs of the target-domain training set except the 100,000 sentence pairs into two equal portions.
We then used the first half and the second half as simulated monolingual data for the source language and the target language, respectively. 
}
\shoetsu{
The monolingual data was used for training SentencePiece and \textsc{cbow} vectors in the target domain and data augmentation by back-translation.
When models did not use the monolingual data, the data used for training SentencePiece and \textsc{cbow} vectors was exactly identical to the training set in each domain.
}
\begin{table}[!t]
\footnotesize
\centering
\begin{tabular}{@{\,}l@{\,}r@{\,\,}|@{\,\,}l@{\,}r@{\,}}
\toprule
   \# encoder/decoder layers & 6    & Label smoothing rate & 0.1 \\ 
   \# attention heads & 8    & Init. learning rate & 1e-3 \\
   Dim. of embeddings      & 512    &  \hfill (warmup) & 1e-7 \\
   Dim. of Transformer     & 2048    & Dropout rate & 0.1 \\ 
   Vocab. size (enc\&dec)  & 16k  & Beam size for decoding & 5 \\
   Max. tokens in batch   & 64k    & Length penalty & 1.2 \\
   
\bottomrule
\end{tabular}
\caption{Hyperparameters of \textsc{nmt} models.} 
\label{tab:hyperparameters}
\end{table}

\subsection{Models and Embeddings}

We adopted Transformer-base~\cite{vaswani2017attention} implemented in fairseq (v0.8.0)\footnote{\url{https://github.com/pytorch/fairseq}}~\cite{ott-etal-2019-fairseq}, as the core architecture for the \textsc{nmt} models.\footnote{Note that since Transformer shares the embedding and output layers of the decoder, vocabulary adaptation is applied to the embedding layer of the encoder and the tied embedding/output layer of the decoder, respectively.}
Major hyperparameters are shown in \tabref{hyperparameters}.\footnote{\shoetsu{For De$\rightarrow$En translation, we made minor modifications to the architecture to follow \citet{hu-etal-2019-domain-adaptation}. 
Concretely, we added layer normalization~\cite{ba2016layer} before each of the encoder and decoder stacks. We also applied dropout to the outputs of the activation functions and self-attention layers. 
}}
We evaluated the performance of the models on the basis of \textsc{bleu} ~\cite{papineni-etal-2002-bleu}.
Before pre-training the models, we induced subword embeddings from the monolingual corpus by Continuous Bag-of-Words (\textsc{cbow})~\cite{mikolov2013distributed} to initialize the embedding layers of the \textsc{nmt} models.

To evaluate the effect of vocabulary adaptation, 
we compared the following settings (and their combinations) that used either or both the source- and target-domain parallel data. 

\smallskip\noindent\textbf{Out-/In-domain}
        trains a model only from the training set in the source/target domain.
        
\smallskip\noindent\textbf{Fine-tuning w/ source-domain vocab.~(FT-srcV)}
continues to train the \textbf{Out-domain} model using the training set in the target domain without any vocabulary adaptation~\cite{luong2015stanford}.
        
\smallskip\noindent\textbf{Fine-tuning w/ target-domain vocab.~(FT-tgtV)}
    Refer to \ssecref{ft-tgtv}. 
    
\smallskip\noindent\textbf{Multi-domain learning (MDL)}
        trains a model from both source and target domain training sets.
        We employed domain token mixing~\cite{britz2017effective} as a 
        method of multi-domain learning.
        In this setting, we jointly used the source and target domain training sets for training subword tokenization models, \textsc{cbow} vectors, and training \textsc{nmt} models (e.g., 2797k + 100k for En$\rightarrow$Ja translation). 

\smallskip\noindent\textbf{Vocabulary Adaptation (VA)}
        Refer to \ssecref{vocabulary-adaptation}.
        We compared two projection methods:
        linear orthogonal transformation (\textbf{VA-Linear}, \S~\ref{subsec:linearmap}) 
        and locally linear mapping (\textbf{VA-LLM}, \S~\ref{subsec:LLM}).
        For \textbf{VA-LLM}, the number of nearest neighbors, $k$, was fixed to 10.\footnote{We evaluated \textbf{VA-LLM} with k=\{1, 5, 10, 20\}, and the default value (k=10) was the best.}
        \shoetsu{To highlight the importance of embedding projection for the proposed method, we also evaluated settings using the target-domain \textsc{cbow} vectors for the re-initialization as is (VA-CBoW).}
    
\smallskip\noindent\textbf{Back-translation (BT)} applies a backward translation to target-domain monolingual data in the target language. 
We employed the most standard back-translation proposed by~\citet{sennrich-etal-2016-improving}.
For this back-translation, 
a backward model (\textit{e.g.}, Ja $\rightarrow$ En) is independently trained from the source-domain parallel data with the same setting and data as \textbf{Out-domain}. The subsequent fine-tuning is applied with the generated pseudo-parallel target-domain corpora and a target-domain training set.

\begin{table}[!t]
\footnotesize
\centering
\begin{tabular}{lrrrr}
\toprule
& \multicolumn{4}{c}{\textbf{\# In-domain data}} \\ 
& \multicolumn{2}{c}{\textbf{En $\rightarrow$ Ja}} & \multicolumn{2}{c}{\textbf{De $\rightarrow$ En}} \\
\cmidrule(lr){2-3}
\cmidrule(lr){4-5}
    & 100k & 2000k & 100k & 715k \\
  \midrule
  \textit{No adaptation}\\
  \,\,\textbf{Out-domain}  &  \multicolumn{2}{c}{4.61} & \multicolumn{2}{c}{2.58} \\
  \,\,\textbf{In-domain}   &  11.69 & 41.83 & 18.79 & 34.16 \\
  \midrule
  \textit{Baselines} \\ 
  \,\,\textbf{MDL}           & 21.65 & 41.92 & 24.03 & 37.74 \\
  \,\,\textbf{FT-srcV}       & 21.45 & 43.09 & 24.59 & 38.43 \\
  \,\,\textbf{FT-tgtV}       & \textbf{28.08} & 42.32 & 24.87 & 36.38  \\
  \midrule
  \textit{Proposed} \\
  \,\,\textbf{VA-CBoW}      & 15.28  & 41.44 & 21.88 &  36.34 \\
  \,\,\textbf{VA-Linear}      & 22.26 & 42.70 & 25.20 & 37.00 \\
  \,\,\textbf{VA-LLM}       & 21.79 & \textbf{43.96} & \textbf{26.40} & \textbf{39.41} \\ 
  \bottomrule
\end{tabular}
\caption{Case-sensitive \textsc{bleu} scores for \textsc{nmt} domain adaptation: En$\rightarrow$Ja  from \textbf{JESC} to \textbf{ASPEC} and De$\rightarrow$En from \textbf{IT} to \textbf{Law}. Size of training set for \textbf{Out-domain} was 2797k for \textbf{JESC} and 338k for \textbf{Law}.
\label{tab:main-results-trans-deenenja}}
\end{table}

Among the above methods, \textbf{Out-domain} and \textbf{In-domain} do not perform domain adaptation.
\textbf{FT-srcV}, \textbf{FT-tgtV}, 
and \textbf{MDL}
are baseline domain adaptation methods. 
\textbf{BT} is 
\yn{applied to \textbf{FT-srcV}, \textbf{FT-tgtV}, and \textbf{VA} for data augmentation.}

Note that 
\textbf{FT-tgtV} and \textbf{MDL} assume that the target domain is given before training with the source-domain data. Although this assumption enables us to build a suitable vocabulary for the target domain, 
it sacrifices the domain portability of trained models. As a result, it requires us to perform training for a long period of each combination of a source and a target domain.

We used Adam~\cite{kingma2014adam} to train each model with the above settings.
During both pre-training and fine-tuning, the learning rate linearly increased for warm-up for the first 4,000 training steps and then decayed proportionally to the inverse square root of the number of updates.
Prior to fine-tuning, we reset the optimizer and the learning rate and then continued training on the training set in the target domain.

\begin{table}[!t]
\footnotesize
\centering
\begin{tabular}{l@{\quad}c@{\quad}c@{\quad}c@{\quad}c@{\quad}c@{\quad}c}
\toprule

    
    & \textbf{Enc} & \textbf{Dec} & \multicolumn{2}{c}{\textbf{En $\rightarrow$ Ja}} & \multicolumn{2}{c}{\textbf{De $\rightarrow$ En}} \\
    \cmidrule(l.{0pt}r.{0pt}){4-5}
\cmidrule(l.{0pt}r){6-7}
    & & & 100k & 2000k & 100k & 715k \\

\midrule
\textbf{FT-srcV} & & & 21.45 & 43.09 & 24.59 & 38.43 \\
\midrule
   & \checkmark &  & \textbf{22.69} & 43.48 & 25.64 & 39.48 \\
   \textbf{VA-LLM} &            & \checkmark                       & 20.75 & 43.66 & 25.69 & \textbf{40.19} \\
         & \checkmark & \checkmark                       &  21.79 & \textbf{43.96} & \textbf{26.40} & 39.41 \\

  \bottomrule
\end{tabular}
\caption{\textsc{bleu} scores on ablation tests for \textbf{VA-LLM}. 
\label{tab:ablation-test}}
\end{table}

\section{Results}\label{sec:results}
\subsection{BLEU Scores}
\tabref{main-results-trans-deenenja} shows the results for the 
domain adaptations. 
Among all the methods, \textbf{VA-LLM} achieved
the best \textsc{bleu} score in three out of the four cases. The low \textsc{bleu} scores for \textbf{Out-domain} show how much domain mismatch degraded the \textsc{nmt} performance, as pointed out in \cite{koehn-2017-six}. 
There were large differences in the performance among
\textbf{VA-*} models that perform vocabulary adaptation prior to
fine tuning. The results confirmed that not only the differences in
the vocabulary (set of subwords) but also the initial embeddings matter in fine-tuning \textsc{nmt} models. 

\textbf{VA-*} methods did not work well in En$\rightarrow$Ja translation when only the 100k target-domain
parallel data was used.
This is probably because the more noisy emebeddings (ambiguous subwords) introduced by the large number of domain-specific words in the ASPEC dataset (\tabref{data-stat}) hinders the embedding projection of \textbf{VA-LLM} and \textbf{VA-Linear} with low-quality \textsc{cbow} vectors trained from the 100k sentences.
In this setting, we need more parallel data for fine-tuning to adjust the noisy initial embeddings.

\tabref{ablation-test} shows results of ablation tests to examine for which side (encoder or decoder) \textbf{VA-LLM} benefited. The results confirmed that the poor performance in En$\rightarrow$Ja translation with the 100k target-domain parallel data is due to the failure of handling semantic shifts in the decoder.\footnote{\yn{We observed the same tendency when we conducted the ablation tests for Ja$\rightarrow$En translation with the ASPEC datasets.}} 
 
The improvements obtained by \textbf{VA-Linear} were modest
overall. This was due to the nature of the linear projection
employed for cross-domain embedding mapping as discussed in
\ssecref{LLM}. We analyze the difference between the two types of
projected embeddings in \ssecref{analysis-emb}.

\begin{table}[!t]
\footnotesize
\centering
\begin{tabular}{lcccc}
\toprule
& \multicolumn{4}{c}{\textbf{\# In-domain data}} \\ 
& \multicolumn{2}{c}{\textbf{En $\rightarrow$ Ja} } & \multicolumn{2}{c}{\textbf{De $\rightarrow$ En}} \\
\cmidrule(lr){2-3}
\cmidrule(lr){4-5}
    & 100k & \textbf{+BT}  & 100k & \textbf{+BT} \\
    
\midrule
\,\,\textbf{FT-srcV}   &  21.45 & 24.63 & 24.59 & 25.81 \\   
\midrule

\multicolumn{5}{l}{\textit{w/ monolingual data for training CBoW}}\\

\,\,\textbf{FT-tgtV}   &  18.85 & 21.75 & 21.87 & 24.49\\
\,\,\textbf{VA-Linear} &  19.35 & 22.19 & 24.09 & 25.79 \\   
\,\,\textbf{VA-LLM}    &  \textbf{25.31} & \textbf{29.73} & \textbf{27.87} & \textbf{28.43} \\    
\bottomrule
\end{tabular}
\caption{Case-sensitive BLEU scores when employing target-domain monolingual data (950k for En$\rightarrow$Ja and 308k for De$\rightarrow$En). 
+BT indicates that monolingual data was used also for data augmentation.
\label{tab:monolingual}}
\end{table}


\subsection{Effects of Monolingual Data}
\tabref{monolingual} shows how employing target-domain monolingual data affected domain adaptation. 
In the settings, the SentencePiece and \textsc{cbow} vectors of the target domain were trained from both the 100k parallel data and the monolingual data (950k and 308k for En$\rightarrow$Ja and De$\rightarrow$En, respectively).
We also evaluated the orthogonality of the proposed method to 
\textbf{BT}
since both methods exploit target-domain monolingual data. 

Interestingly, the results of \textbf{FT-tgtV} and \textbf{VA-Linear} were worse than 
the results
in \tabref{main-results-trans-deenenja}.
We consider the reason to be as follows. 
When additionally using the target-domain monolingual data, 
\yn{the resulting SentencePiece model and \textsc{cbow} vectors become more suitable for the target domain
thanks to the increase of data.}
However, this also means that target-domain-specific words appearing only in the monolingual data 
accelerated
the vocabulary mismatches, the semantic shifts, and the difference of topology in the embedding space. 
As the result, the vocabulary mismatches degraded the pre-trained model of the source domain for \textbf{FT-tgtV} and 
linear transformation failed to handle the semantic shifts for \textbf{VA-Linear}.


In contrast, due to the capability of the projection method, the performance of \textbf{VA-LLM} was successfully improved by the use of the monolingual data.
\tabref{monolingual} also shows  
the orthogonality of \textbf{VA-LLM} to \textbf{BT}, since
the increase of \textsc{bleu} scores for \textbf{VA-LLM + BT} from \textbf{FT-srcV + BT} were substantial
(5.10 pt and 2.61 pt for En$\rightarrow$Ja and De$\rightarrow$En translation, respectively).

\begin{table}[!t]
\footnotesize
\centering
\begin{tabular}{lccc}
\toprule
   & \multicolumn{3}{c}{\textbf{\# Updates in training w/}} \\
         & \textbf{source} & \textbf{target} & \textbf{BT} \\
         & (2797k) & (100k) & (950k) \\
\midrule
  \textit{w/o monolingual data} \\
  \quad\textbf{In-domain}          & -       & 3,440 & -\\
  \quad\textbf{MDL}         & \multicolumn{2}{c}{36,342} & - \\
  \quad\textbf{FT-srcV}          & 28,750  & 2,480 & -\\
  \quad\textbf{VA-LLM}     & 28,750 & 3,200 & - \\
  
  \textit{w/ monolingual data} \\
  \quad\textbf{FT-srcV + BT}          & 56,350 & \multicolumn{2}{c}{31,280}  \\
  \quad\textbf{VA-LLM + BT} & 56,350 & \multicolumn{2}{c}{32,895} \\
  
  \bottomrule
\end{tabular}
\caption{Number of updates until convergence for En$\rightarrow$Ja translation. 
\label{tab:convergence-steps}}
\end{table}


  

\begin{CJK}{UTF8}{min}
  \begin{table*}[t]
\small
    \centering
    \begin{tabular}{lp{12cm}}
    \toprule
    \textbf{Input (JESC vocab.)} & \_3 \_cases \_of \_the \_lu m bar \underline{\textcolor{red}{\_spinal}}$_1$ \_can al \underline{\textbf{\textcolor{red}{\_ste no s is}}}$_2$ $\cdots$ \\
    \textbf{Input (ASPEC vocab.)} & \_3 \_cases \_of \_the \_lumbar \underline{\textcolor{red}{\_spinal}}$_1$ \_canal \underline{\textbf{\textcolor{red}{\_stenosis}}}$_2$ $\cdots$ \\
    \midrule
    \textbf{Reference} & $\cdots$ 腰部 \underline{\textcolor{red}{\textbf{脊柱}}}$_1$ 管 \underline{\textcolor{red}{\textbf{狭窄} 症}}$_2$ の 3 例 に つ い て $\cdots$ \\
    \midrule
    \textbf{FT-srcV} & $\cdots$ 腰部 \verb|<unk>| 柱 管 狭 \verb|<unk>| 症 の 3 症例 に つ い て $\cdots$ \\
    \textbf{FT-srcV + BT} & $\cdots$ 腰部 \verb|<unk>| 柱 管 狭 \verb|<unk>| 症 の 3 症例 に つ い て $\cdots$ \\
    \textbf{VA-LLM + BT} & $\cdots$ 腰部 \underline{\textbf{\textcolor{red}{脊柱}}}$_1$ 管 \underline{\textbf{\textcolor{red}{狭窄}}}$_2$ の 3 症例 に つ い て $\cdots$ \\
    
%
    \bottomrule
    \addlinespace
    \toprule
    \textbf{Input (IT vocab.)} &  \_falls \_der \_Austausch \_der \underline{\textbf{\textcolor{red}{\_Rat if ik ation s ur ku nden}}}$_1$ \_zwischen $\cdots$\\
    \textbf{Input (Law vocab.)} & \_falls \_der \_Austausch \_der \underline{\textbf{\textcolor{red}{\_Ratifikation surkunde n}}}$_1$ \_zwischen $\cdots$ \\
    \midrule
    \textbf{Reference} & should the \underline{\textcolor{red}{instruments of \textbf{ratification}}}$_1$ be exchanged between $\cdots$ \\
    \midrule
    \textbf{FT-srcV} & if the exchange of the \textcolor{red}{\textbf{ratification}} of \textcolor{red}{\textbf{ratification}} between $\cdots$ \\
    \textbf{FT-srcV + BT} & where the exchange of the Council takes place between $\cdots$ \\
    \textbf{VA-LLM + BT} & if the \underline{\textcolor{red}{instruments of \textbf{ratification}}}$_1$ are met between $\cdots$ \\
    \bottomrule

    \end{tabular}
    \caption{Translation examples of the 
    models with 100k target-domain parallel data in 
    \tabref{main-results-trans-deenenja} and 
    \tabref{monolingual}.
    \textbf{Bolded words}
    are rare or unknown in source domain. 
    \underline{Underlined words} and subscript numbers indicate correspondence. 
    Input (JESC, IT) and Input (ASPEC, Law) were fed to \textbf{FT-srcV}/\textbf{FT-srcV + BT} and \textbf{VA-LLM + BT}, respectively.
    \label{tab:output-examples-trans}}
\end{table*}
\end{CJK}

\subsection{On Efficiency: Training Steps}
\tabref{convergence-steps} shows the number of updates until convergence in En$\rightarrow$Ja translation with the 100,000 target-domain training set.\footnote{As for \textbf{FT-srcV + BT} and \textbf{VA-LLM + BT}, the number of updates in the pre-training phase is the sum of the training steps for both forward and backward models.}
We confirmed that all models were trained over a sufficient number of steps. The validation loss did not improve over at least five epochs after the best model was chosen.
We used four GPUs (NVIDIA Quadro P6000) for training, and it took 0.9 sec/update on average.

Here, we emphasize that \textbf{VA-LLM} achieved superior performance with a small number of updates 
(3,200 steps, less than 50 minutes) 
similarly to \textbf{FT-srcV}.
Note that the overhead time of our vocabulary adaptation was negligible since embedding projection took only several minutes.
Meanwhile, \textbf{FT-srcV + BT} took 31,280 steps due to the size of the augmented data even when we ignore the time taken to generate back-translated parallel data.

Additionally, our proposed method is based on fine-tuning and the target domain is not supposed to be given before pre-training in the source domain, differently from \textbf{MDL}. 
Therefore, the pre-trained \textbf{Out-domain} can be reused each time when the target domain or settings are changed, which enables us to omit the long training time (28,750 steps, about 7.2 hours) per model training.
As the training steps of \textbf{VA-LLM + BT} show, the overhead  caused by employing the proposed method with back-translation was also small. Nevertheless, the improvements of \textbf{VA-LLM + BT} compared with \textbf{FT-srcV + BT} were substantial (\tabref{monolingual}).


\section{Analysis}
\subsection{Translation Examples}
\begin{CJK}{UTF8}{min}
\tabref{output-examples-trans} shows translation examples generated by 
\textbf{FT-srcV} in \tabref{main-results-trans-deenenja}, \textbf{FT-srcV + BT} and \textbf{VA-LLM + BT} in \tabref{monolingual}. 
The size of target-domain parallel data for training was 100k.

\textbf{FT-srcV} and \textbf{FT-srcV + BT}
often failed to translate target-domain-specific words that were \yn{tokenized} into 
short subwords.
In such cases,
\yn{the} 
models tended to ignore or transliterate them.
For instance, the De$\rightarrow$En examples (lower) show that 
\textbf{FT-srcV} and \textbf{FT-srcV + BT} failed in translating ``Ratifikationsurkunden (instruments of ratification).''

Moreover, in the En$\rightarrow$Ja examples (upper), the decomposed target-domain-specific words ``脊柱 (spinal)'' and ``狭窄症 (stenosis)'' contained target-domain-specific subwords such as ``脊'' and ``窄.'' 
The models without vocabulary adaptation also failed to handle these subwords when both the source-domain training set and the target-domain 100k training set rarely contained them.

Meanwhile, \textbf{VA-LLM + BT} successfully translated both of the cases with the help of target-domain monolingual data.
These examples imply the difficulty in translating target-domain-specific words without vocabulary adaptation.

We observed that 
\textbf{VA-LLM + BT} 
generated
various target-domain-specific words. 
To quantitatively
confirm this, we calculated the percentage of distinct words included in both the generated outputs and the references.
The outputs in En$\rightarrow$Ja translation generated by \textbf{VA-LLM + BT}, \textbf{FT-srcV + BT}, and \textbf{FT-srcV} contained 57.9\%, 53.4\%, and 49.5\% of distinct words in the references, respectively. 
\end{CJK}

\begin{figure}[t]
\centering
\includegraphics[width=\linewidth,clip]{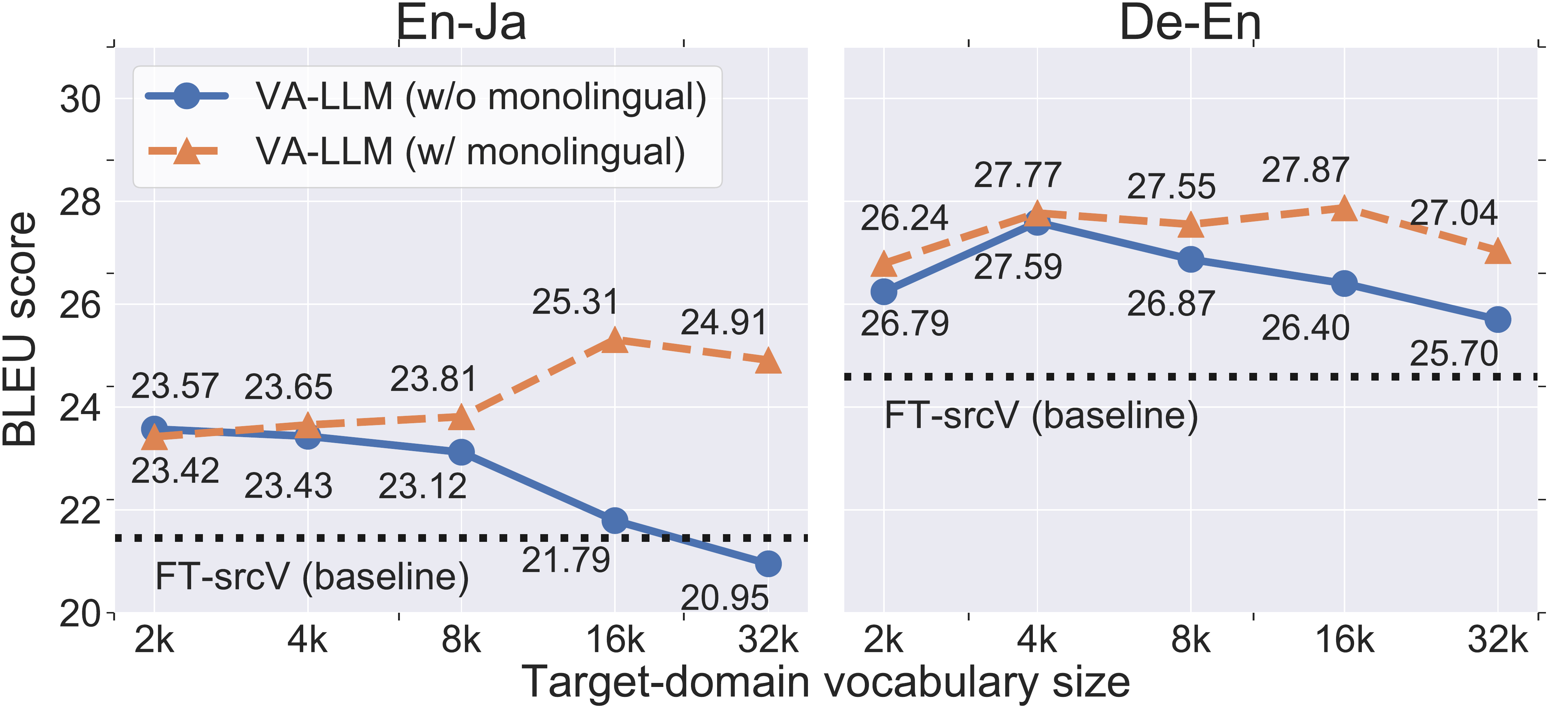}
\caption{BLEU scores of \textsc{va-llm} while varying 
target-domain vocabulary size. The source-domain vocabulary size was fixed to 16k.}
\label{fig:vocabulary-size}
\end{figure}

\subsection{Effect of Vocabulary Size \yn{in Fine Tuning}}
As reported in \cite{sennrich-zhang-2019-revisiting},
the vocabulary size of an \textsc{nmt} model can affect its translation quality
in a low-resource setting.
How about in fine-tuning? To explore this, we varied only the \textbf{target-domain} vocabulary size of \textbf{VA-LLM} before fine-tuning by vocabulary adaptation. 

\figref{vocabulary-size} shows that \textbf{VA-LLM} preferred large vocabulary sizes when \yn{additional} target-domain monolingual data was used \yn{for training \textsc{cbow}}, whereas it preferred small vocabulary sizes when the data 
was not used.
We consider the reason to be as follows. In the former case, a large vocabulary 
contains
low-frequency subwords of which representation is unlikely to be well-trained as discussed in \cite{sennrich-zhang-2019-revisiting}. 
In the latter case, however, target-domain monolingual data can cover such low-frequency subwords. 

As this analysis showed, the vocabulary size also had large effects on fine-tuning (3.52 pt difference at most).
Besides the vocabulary mismatch problem, our vocabulary adaptation could make further improvements by the vocabulary size were adjusted depending on the amount of target-domain parallel and monolingual data with a low training cost. 





\begin{table}[t]
\footnotesize
\centering
\begin{tabular}{@{\,}l@{\,\,}l@{\,}}
\toprule
\multicolumn{2}{@{}l}{\textbf{Nearest neighbors in ASPEC-\textsc{cbow} embedding space}}\\
\midrule
    \_branches       & \_branch, \_roots, \_veins, \_arteries, \_trees \\
                  
    \_experimentally & \_systematically, \_numerical, \_theoretical, \\  
                     & \_experimental, \_experiments \\
\toprule
\multicolumn{2}{@{}l}{\textbf{Nearest neighbors in JESC-\textsc{nmt} embedding space}}\\
\midrule
\multicolumn{2}{@{}l}{\textit{via linear transformation (Linear)}} \\
    \_branches           & \textbf{\_trees}, \_sides, \_birds, \_parts, \_pieces \\
    \_experimentally     & \_rope, \_tanks, \_laser, \\
                         & \_gravitational, \_simulation \\
    \midrule
    \multicolumn{2}{@{}l}{\textit{via locally linear mapping (LLM)}} \\
    \_branches       & \textbf{\_branch}, \textbf{\_trees}, \textbf{\_roots}, \textbf{\_veins}, \textbf{\_arteries} \\
    \_experimentally & \_by, \_experiment, \textbf{\_experiments}, \\ 
                     & \textbf{\_experimental}, \_simulation  \\
\bottomrule
\end{tabular}

\caption{Top-5 nearest neighbors of 
``\textit{\_branches}'' and ``\textit{\_experimentally}'' 
in ASPEC-\textsc{cbow} embedding space and JESC-\textsc{nmt} embedding space via cross-domain embedding projection: \textbf{bold-faced} subwords are nearest neighbors shared across both top-5. The ASPEC-\textsc{cbow} vectors are trained from the 100k target-domain parallel data and the monolingual data.
\label{tab:llm-example}}
\end{table}

\subsection{Quality of Cross-domain Embeddings}\label{subsec:analysis-emb}

The advantage of our approach is that it adjusts the meanings of subwords (embeddings) as well as the vocabulary (set of subwords) to the target domain.
We thus examined to what extent our vocabulary adaptation captures the semantic shift. 

We first observed the nearest neighbors based on cosine similarity for each of the subword embeddings in the target domain (hereafter, ASPEC-\textsc{cbow}).\footnote{Through this analysis, the candidates of nearest neighbors were limited to 
 the shared subwords across JESC and ASPEC domains for clear comparison.}
Note that the nearest neighbors should be unchanged even after embedding projection to keep the meanings learned in the target domain.

Next, we compute cosine similarities between each of the projected  ASPEC-\textsc{cbow} and the embeddings of \textbf{Out-domain} to find their nearest neighbors in the embedding space of \textbf{Out-domain} (hereafter, JESC-\textsc{nmt}). 
The obtained nearest neighbors show how the ASPEC-\textsc{cbow} embeddings projected by linear-transformation or LLM performed during fine-tuning.






 \tabref{llm-example} shows the nearest neighbors of two words: ``\textit{\_branches},'' which appears in both domains and can have different meanings across domains, and ``\textit{\_experimentally},'' 
which is only in the ASPEC domain.

While the \textsc{cbow} vector for ``\textit{\_branches}'' and the embedding projected by LLM have the meaning of ``\_veins'' and ``\_arteries'', the embedding projected by linear transformation lost it.   ``\textit{\_experimentally}'' 
 is a subword that only the target-domain (ASPEC) vocabulary contains.
 As illustrated in \figref{cross-domain-mapping}, the mapping of target-domain-specific subword embeddings is likely to fail due to the difference of topology in the embedding space.
 We found that LLM relatively accurately computed its embedding in the JESC-\textsc{nmt} space while linear transformation failed.
 This tendency was also observed when using only the 100k parallel data for training of SentencePiece and \textsc{cbow} vectors.
 These observations demonstrate the capability of LLM in cross-task/domain embedding projection.
 


\section{Conclusion}\label{sec:conclusion}
In this study, we tackled 
the vocabulary mismatch problem in domain adaptation for \textsc{nmt}, and we proposed vocabulary adaptation, a simple but
\yn{direct solution to this problem}. It 
adapts the vocabulary of a pre-trained \textsc{nmt} model to a target domain for performing effective fine-tuning. Regarding domains as independent languages/tasks, our method  
makes wide-coverage word embeddings induced from target-domain monolingual data be compatible with a model pre-trained in a source domain. 

We explored two methods for projecting word embeddings across two domains: linear transformation and locally linear mapping (LLM).
The experimental results for English to Japanese translation and German to English translation
confirmed that our domain adaptation method with LLM dramatically improved the translation performance.

Although the vocabulary adaptation was evaluated only for \textsc{nmt}, it is
also applicable to a wider range of neural network models and tasks, and it can even be combined with existing fine-tuning-based domain adaptations.
We will release all 
code to promote the reproducibility of our results.\footnote{\url{{https://github.com/jack-and-rozz/vocabulary_adaptation}}}

\section*{Acknowledgements}
This work was partially supported by JSPS KAKENHI Grant Numbers 19J14522 and 20J13810.
The research was partially supported by NII CRIS collaborative research program operated by NII CRIS and LINE Corporation.

\bibliography{emnlp2020}
\bibliographystyle{acl_natbib}
 
\end{document}